\documentclass[conference]{IEEEtran}
\IEEEoverridecommandlockouts
\usepackage{cite}
\usepackage{amsmath,amssymb,amsfonts}
\usepackage{algpseudocode}
\usepackage{algorithm}
\usepackage{booktabs}
\usepackage{balance}
\usepackage{graphicx}
\usepackage{multirow}
\usepackage{textcomp}
\usepackage{colortbl}
\usepackage{placeins}
\usepackage{tikz}
\usetikzlibrary{arrows.meta,positioning,calc,fit}
\usepackage{xcolor}
\usepackage[hidelinks]{hyperref}
\def\BibTeX{{\rm B\kern-.05em{\sc i\kern-.025em b}\kern-.08em
    T\kern-.1667em\lower.7ex\hbox{E}\kern-.125emX}}
\begin{document}

\title{T2ST: Data-Efficient Spatio-Temporal Foundation Models via Temporal Data and Knowledge Transfer}

\author{
\IEEEauthorblockN{\makebox[0.30\textwidth][c]{1\textsuperscript{st} Yue Xu}}
\IEEEauthorblockA{\textit{Faculty of Information Science} \\
\textit{and Computing} \\
\textit{University of Macau}\\
Macau, China \\
yc57466@um.edu.mo}
\and
\IEEEauthorblockN{\makebox[0.30\textwidth][c]{2\textsuperscript{nd} Kaiyun Wei}}
\IEEEauthorblockA{\textit{Faculty of Information Science} \\
\textit{and Computing} \\
\textit{University of Macau}\\
Macau, China \\
yc57929@um.edu.mo}
\and
\IEEEauthorblockN{\makebox[0.30\textwidth][c]{3\textsuperscript{rd} Wenying Duan}}
\IEEEauthorblockA{\textit{School of Artificial} \\
\textit{Intelligence} \\
\textit{Nanchang University}\\
Nanchang, China \\
wenyingduan@ncu.edu.cn}
\and[\hfill\mbox{}\par\vspace{1em}\mbox{}\hfill]
\IEEEauthorblockN{\makebox[0.30\textwidth][c]{4\textsuperscript{th} Zichen Xu}}
\IEEEauthorblockA{\textit{School of Computer Science} \\
\textit{Intelligence} \\
\textit{Nanchang University}\\
Nanchang, China \\
xuz@ncu.edu.cn‌}
\and
\IEEEauthorblockN{\makebox[0.30\textwidth][c]{5\textsuperscript{th} Shuai Ma}}
\IEEEauthorblockA{\textit{School of Computer Science} \\
\textit{and Engineering} \\
\textit{Beihang University}\\
Beijing, China\\
mashuai@buaa.edu.cn}
\and
\IEEEauthorblockN{\makebox[0.30\textwidth][c]{6\textsuperscript{th} Xiaoxi He}}
\IEEEauthorblockA{\textit{Faculty of Information Science} \\
\textit{and Computing} \\
\textit{University of Macau}\\
Macau, China \\
hexiaoxi@um.edu.mo}
}

\maketitle

\begin{abstract}
Spatio-temporal foundation models (STFMs) require large-scale training corpora, yet synchronized observations and spatial metadata are costly to acquire and curate, limiting the availability of suitable data. By contrast, time-series data are abundant and easier to collect. Since observations at each location form a time series, this motivates a natural question: can abundant time-series data enhance STFMs when spatio-temporal observations are scarce? To bridge this gap, we propose T2ST, a data-efficient training framework that augments limited spatio-temporal observations with selected auxiliary time series and transfers knowledge from a frozen time-series foundation model. Our key insight is that temporal pattern learning can benefit from abundant external time series, while spatial dependency learning remains grounded in real spatio-temporal observations. Guided by this insight, T2ST retrieves matched sequences to reinforce sparsely observed temporal patterns and selects reliable complementary sequences to broaden temporal coverage. Temporal relation distillation aligns relationships among historical patches, while multi-scale prediction distillation transfers forecasting behavior at fine and coarse temporal resolutions, enabling knowledge transfer across heterogeneous architectures. Using only 10\% of the original spatio-temporal data, T2ST enables STFMs to achieve overall forecasting performance competitive with or superior to their full-data counterparts across six long-term and twelve short-term forecasting datasets. These results establish a new paradigm for data-efficient training of spatio-temporal foundation models. Code is available at https://anonymous.4open.science/r/T2ST-65F3/.
\end{abstract}


\section{Introduction}
Spatio-temporal foundation models (STFMs) aim to support forecasting across diverse application domains, including intelligent transportation~\cite{li2017diffusion,ma2024learning,fang2023spatio,han2025scalable}, energy management~\cite{zhou2022sdwpf}, and environmental monitoring~\cite{zheng2013u}. By pre-training on heterogeneous spatio-temporal datasets, these models move beyond the conventional one-model-per-dataset paradigm and enable knowledge sharing across cities and sensing systems~\cite{li2025open,yuan2024unist,han2025scalable,chen2026urbanfm}. However, the scarcity of spatio-temporal training data remains a major bottleneck for STFMs. Building large-scale corpora involves collecting observations across multiple locations and curating both the measurements and their associated spatial information. The resulting acquisition and curation costs make these corpora difficult to expand. This scarcity motivates data-efficient STFM training: maintaining forecasting performance with fewer spatio-temporal observations. This scarcity motivates \textbf{\emph{data-efficient STFM training}}: maintaining forecasting performance with fewer spatio-temporal observations.

In contrast, time-series data can be collected from diverse sources without requiring explicit spatial information~\cite{lai2023lightcts}. This flexibility has supported substantially larger pre-training corpora for time-series foundation models (TSFMs)~\cite{pmlr-v235-goswami24a,pmlr-v235-liu24cb}, such as Moirai, Chronos, and TimesFM~\cite{woo2024unified,ansari2025chronos,das2023decoder}. As summarized in Table~\ref{tab:pretraining-data-scale}, reported TSFM corpora contain tens to hundreds of billions of time points, compared with hundreds of millions to over a billion for representative STFMs. Pre-training on these large and diverse corpora enables TSFMs to learn temporal regularities, including trends, periodicities, and dependencies across time scales. These regularities are also relevant to spatio-temporal forecasting, since observations at each location form a time series. 
We therefore ask: \textbf{\emph{Can abundant time-series data and pretrained temporal knowledge reduce the amount of structured spatio-temporal data needed to train effective STFMs?}} To investigate this question, we supplement limited spatio-temporal training data with auxiliary time series and transfer temporal knowledge from a pretrained TSFM to the STFM.

\begin{table}[t]
    \centering
    \caption{Pre-training data scales of representative spatio-temporal and time-series foundation models. The B denotes $10^9$ reported observations or time points.}
    \label{tab:pretraining-data-scale}
    \begingroup
    \small
    \setlength{\tabcolsep}{3pt}
    \renewcommand{\arraystretch}{1.15}
    \begin{tabular*}{\linewidth}{@{\extracolsep{\fill}}lr@{}}
        \noalign{\hrule height 1pt}
        \noalign{\smallskip}
        \textbf{Model} & \textbf{Points (B)} \\
        \noalign{\smallskip}
        \hline
        \noalign{\smallskip}
        \multicolumn{2}{@{}l}{\textit{Spatio-temporal foundation models (STFMs)}} \\
        \noalign{\smallskip}
        OpenCity~\cite{li2025open} & $\approx 0.15$ \\
        CompactST~\cite{han2025scalable} & $0.29$ \\
        WorldST~\cite{chen2026urbanfm} & $> 1$ \\
        \hline
        \noalign{\smallskip}
        \multicolumn{2}{@{}l}{\textit{Time-series foundation models (TSFMs)}} \\
        \noalign{\smallskip}
        Moirai~\cite{woo2024unified} & $\mathbf{> 27}$ \\
        Chronos~\cite{ansari2025chronos} & $\mathbf{\approx 84}$ \\
        TimesFM~\cite{das2023decoder} & $\mathbf{\approx 100}$ \\
        \noalign{\smallskip}
        \noalign{\hrule height 1pt}
    \end{tabular*}
    \par\smallskip
    \endgroup
    \vspace{-4mm}
\end{table}

Effectively using these temporal resources requires addressing two challenges under limited spatial supervision. The first is \emph{data selection}. Auxiliary corpora contain heterogeneous temporal dynamics,  not all of which are equally useful for STFM training. Selecting only similar sequences, however, may restrict temporal coverage and overlook complementary patterns. For a foundation model intended to transfer across domains, similarity alone is therefore an incomplete criterion for selecting useful temporal data. The second is \emph{knowledge transfer}. Applying a pretrained TSFM independently to each location does not explicitly model spatial interactions, while transferring its temporal knowledge into an STFM is complicated by differences in architecture, hidden dimensions, and temporal resolution. This raises the challenge of enriching temporal learning with external resources while keeping spatial learning grounded in the available structured observations.

These challenges motivate the design of \textsc{T2ST}, a data-efficient training framework that integrates auxiliary time-series selection with temporal knowledge distillation from a frozen TSFM into an STFM. Our key insight is that \textbf{selected external time series and pretrained temporal knowledge can enrich an STFM's temporal learning, while real spatio-temporal observations continue to provide supervision for spatial dependencies}. This insight allows temporal learning to draw on resources beyond the limited spatio-temporal corpus, with the available structured observations supporting joint temporal and spatial modeling.

To address the data selection challenge, T2ST uses the limited spatio-temporal observations as a reference to construct two auxiliary subsets in a shared temporal representation space. \emph{Matched sequences} are retrieved according to their similarity to the temporal patterns represented in the available reference, providing additional examples of familiar temporal patterns that are sparsely sampled in the available observations. \emph{Diverse sequences} contribute complementary temporal patterns beyond this limited reference, broadening the coverage available for learning. For the diverse subset, teacher reliability filtering retains candidates that the TSFM can forecast reliably, while temporal pattern discovery identifies coherent groups of additional patterns. This selection strategy balances relevance to the observed dynamics with exposure to a broader range of temporal patterns.

To address the knowledge transfer challenge, T2ST distills temporal knowledge through two complementary objectives. \emph{Temporal relation distillation} aligns relationships among historical patches rather than directly matching hidden features, accommodating differences in teacher and student representations. \emph{Multi-scale prediction distillation} aligns forecasts at fine and coarse temporal resolutions, transferring both local variations and broader temporal trends. During training, auxiliary sequences are processed as separate single-node inputs through the student's shared temporal pathway, while real spatio-temporal samples are processed with their spatial structure to support temporal and spatial learning.

In summary, the contributions of this study are as follows:

\begin{itemize}
    \item We introduce a data-efficient training paradigm for STFMs, in which abundant auxiliary time-series data enrich temporal pattern learning while limited real spatio-temporal observations provide spatial supervision. This broadens the sources of training information available to STFMs beyond structured spatio-temporal corpora.
    \item We propose \textsc{T2ST}, a framework that integrates systematic selection of matched and diverse time series with knowledge distillation from a pretrained TSFM. Temporal relation distillation and multi-scale prediction distillation transfer temporal knowledge across heterogeneous teacher and student architectures.
    
    \item We conduct extensive experiments across different spatio-temporal foundation models and datasets. Using only one-sixth of the full spatio-temporal training set, \textsc{T2ST} achieves performance competitive with or superior to that of counterparts trained on the full dataset, demonstrating its effectiveness, generality, and potential to reduce the structured-data cost of future STFMs.

\end{itemize}

\section{Related Work}

\subsection{Spatio-Temporal Foundation Models}

Existing work on spatio-temporal foundation models broadly follows two directions~\cite{liang2025foundation}. The first adapts pretrained large language models (LLMs) to spatio-temporal prediction. UrbanGPT, for example, combines a spatio-temporal dependency encoder with instruction tuning to integrate spatio-temporal observations into an LLM~\cite{li2024urbangpt}. The second develops models specifically for spatio-temporal data and pretrains them directly on heterogeneous spatio-temporal corpora. UniST combines masked pretraining with knowledge-guided prompting for prediction across urban scenarios~\cite{yuan2024unist}, while OpenCity integrates Transformer-based temporal modeling with graph-based spatial modeling for traffic forecasting across cities~\cite{li2025open}. CompactST develops compact models for scalable multi-domain pretraining~\cite{han2025scalable}, and UrbanFM investigates data and architecture scaling for urban spatio-temporal forecasting~\cite{chen2026urbanfm}. These directly pretrained models demonstrate strong transfer across cities and tasks, but their pretraining still relies on substantial spatio-temporal corpora.

To alleviate this constraint, temporal representations are often strengthened prior to or alongside spatial modeling. STEP pretrains a masked time-series encoder for a downstream spatio-temporal graph neural network, while BigST extracts features from long node-wise histories for traffic forecasting~\cite{shao2022pre,han2024bigst}.  Time Tracker separates channel-wise pretraining from inter-variable modeling; FactoST pretrain temporal backbones before spatial adaptation~\cite{liang2025time,zhong2026learning}.

Building on these efforts, \textsc{T2ST} focuses on strengthening existing STFMs under a limited spatio-temporal data budget. It combines selected external time series with guidance from an independently pretrained TSFM, retaining the STFM backbone during joint training.


\subsection{Knowledge Distillation for Forecasting}

Knowledge distillation (KD) transfers knowledge from a trained teacher to a student~\cite{hinton2015distilling,luo2025towards,romero2015fitnetshintsdeepnets}. KD methods are broadly categorized as response-based, feature-based, and relation-based, aligning predictions, intermediate representations, and relationships among representations, respectively~\cite{gou2021knowledge}.Relation-based methods transfer structural knowledge without directly aligning representation spaces~\cite{tung2019similarity,park2019relational,wang2021minilmv2}.


Recent efforts have explored knowledge distillation for time-series and spatio-temporal forecasting~\cite{jin2023transferable,dey2026trust,wang2026s,li2025frequency}. For time-series forecasting, TimeDistill transfers multi-scale and multi-period
patterns from Transformer and CNN based teachers to lightweight MLP students~\cite{ni2026timedistill}. DistilTS further explores distillation from pretrained TSFMs, combining horizon-aware supervision with temporal representation alignment to train compact forecasters~\cite{li2026distilling}. In spatio-temporal forecasting, LightST transfers graph-structured spatial and temporal knowledge from a GNN teacher to an efficient MLP student for traffic prediction~\cite{zhang2025efficient}. These distillation methods primarily address the computational and storage costs of powerful forecasting models by transferring their knowledge to compact students. However, efficient model deployment does not directly address another key challenge in STFM training: the limited availability of observations. 

\textsc{T2ST} bridges this gap by distilling pretrained temporal knowledge into an STFM using abundant temporal-only data, while learning spatial dependencies from the available spatio-temporal observations. This allows temporal learning to draw on resources beyond the limited structured corpus, with the aim of reducing the amount of spatio-temporal data needed to train an effective STFM.


\section{Preliminaries}
\subsection{Spatio-Temporal and Time-Series Data}
Spatio-temporal (ST) data consist of observations collected over time from spatially related entities, such as road sensors or urban regions. Spatial structure is represented by
a graph $\mathcal{G}=(\mathcal{V},\mathcal{E},\mathbf{A})$, where $\mathcal{V}$ contains $N$ spatial entities, $\mathcal{E}$ denotes their relations, and $\mathbf{A}\in\mathbb{R}^{N\times N}$ is the adjacency matrix. The observations are denoted by $\mathbf{S}\in\mathbb{R}^{T\times N\times C}$, where $T$ is the number of timestamps and $C$-dimensional vector indicating $C$ features. Each $\mathbf{S}_{t,:,:}\in\mathbb{R}^{N\times C}$ records the measurements of all entities at time step $t$. For forecasting, we denote the historical observations by $\mathbf{X}_t\in\mathbb{R}^{L\times N\times C}$ and the future targets by $\mathbf{Y}_t\in\mathbb{R}^{H\times N\times C}$, covering time steps $t-L+1$ to $t$ and $t+1$ to $t+H$, respectively. Here, $L$ and $H$ denote the history length and forecasting horizon.

Time-series (TS) data are a time-ordered sequence of observations, denoted by $\mathbf{Z}\in\mathbb{R}^{T\times D}$, where $D$ is the number of observed variables. Each row $\mathbf{z}_t\in\mathbb{R}^{D}$ records the values of
these variables, such as electricity load or temperature, at time step $t$. In particular, each node-wise trajectory $\mathbf{S}_{:,v,:}\in\mathbb{R}^{T\times C}$ extracted from spatio-temporal data is itself a multivariate time series, or a univariate series when $C=1$.

\subsection{Spatio-Temporal and Time-Series Foundation Models}

The STFM is pretrained on observations from multiple spatial domains to learn reusable spatial and temporal patterns. For spatio-temporal forecasting, its prediction is expressed as
\begin{equation}
\label{eq:stfm_prediction}
\widehat{\mathbf{Y}}
= f_{\theta}(\mathbf{X},\mathcal{G})
\end{equation}
where $\theta$ is shared parameters across domains.

The TSFM is pretrained on a large collection of time series to learn transferable temporal patterns. Given an input time series $\mathbf{u}$, its prediction is denoted by
\begin{equation}
\label{eq:tsfm_prediction}
\widehat{\mathbf{v}}
= g_{\phi}(\mathbf{u})
\end{equation}
where $\phi$ denotes the parameters of the TSFM.

\subsection{Problem Formulation}
\label{sec:structured_data_budget}

We aim to train an STFM from limited synchronized observations using auxiliary temporal data and a pretrained TSFM. Across all ST training datasets used in our long-term experiments, eligible windows are selected consecutively from the end of each source's training split, without random sampling across time. Let $K$ be the selected window count and $M$ the total eligible window count over the complete source timelines. The window budget is
\begin{equation}
\label{eq:nominal_st_budget}
r_{\mathrm{win}}=\frac{K}{M}.
\end{equation}
The 10\% setting uses $r_{\mathrm{win}}\approx0.1$.

 For $k>0$ consecutive selected windows from a source, let $L$ and $H$ denote the history and forecast lengths in its native time steps. The union of their input and target timestamps forms a contiguous block containing unique timestamps. 
\begin{equation}
\label{eq:contiguous_st_coverage}
Q=k+L+H-1
\end{equation}

\section{Method}
In this section, we present T2ST, a data-efficient training framework for spatio-temporal foundation models. T2ST augments limited spatio-temporal observations with selected temporal-only data and transfers temporal knowledge from a pretrained time-series foundation model through distillation. We first provide an overview of the framework and and then describe its key components in detail.

\begin{figure*}
    \centering
    \includegraphics[width=0.95\linewidth]{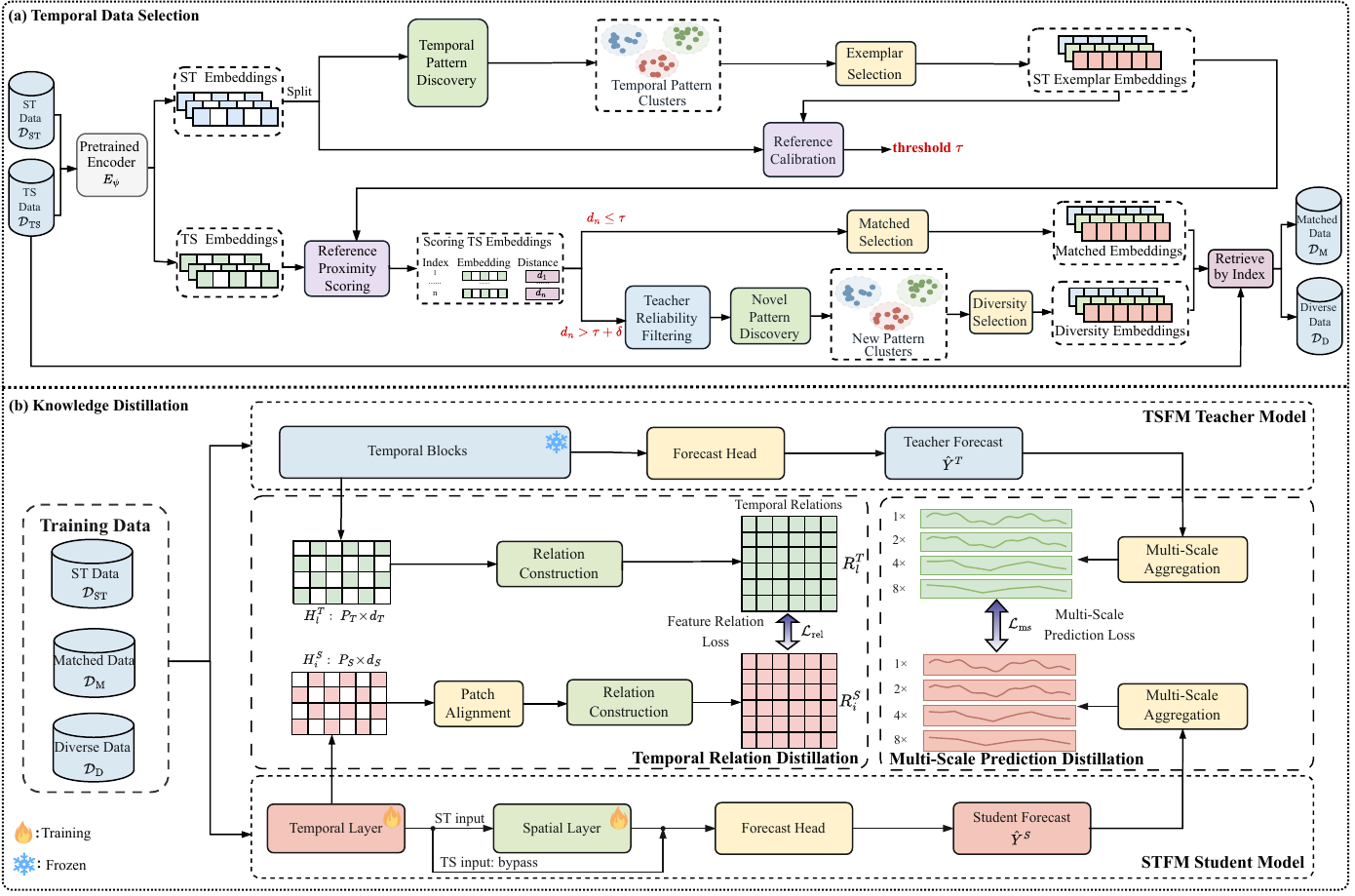}
    \caption{Overview of the T2ST framework. (a) \textbf{Temporal data selection}: a shared encoder maps spatio-temporal and auxiliary time-series windows into a common representation space. A calibrated spatio-temporal reference guides matched selection, while teacher reliability filtering and novel pattern discovery guide diversity selection. Selected embeddings identify the source windows retrieved by index to construct $\mathcal{D}_{\mathrm{M}}$ and $\mathcal{D}_{\mathrm{D}}$. (b) \textbf{Knowledge distillation}: a frozen TSFM guides a STFM student through temporal relation distillation and multi-scale prediction distillation. The student is trained on $\mathcal{D}_{\mathrm{ST}}$, $\mathcal{D}_{\mathrm{M}}$, and $\mathcal{D}_{\mathrm{D}}$, sharing temporal modeling parameters across both input types and bypassing spatial interactions for temporal-only inputs. Snowflakes and flames indicate frozen and trainable parameters, respectively.}
    \label{fig2overview}
\end{figure*}

\subsection{Overall Framework}
As illlustrated in Figure~\ref{fig2overview}, T2ST consists of an offline temporal data selection stage and a distillation training stage. Given limited spatio-temporal observations and an auxiliary time-series corpus, it selects complementary temporal data and transfers knowledge from a pretrained TSFM to an STFM student. The main components are outlined below.

\noindent\textbf{Temporal Data Selection.}
A shared encoder maps node-wise sequences from the spatio-temporal dataset and auxiliary time-series windows into a shared representation space. Using the spatio-temporal data as a reference, we construct a matched subset $\mathcal{D}_{\mathrm{M}}$ containing similar temporal patterns and a diverse subset $\mathcal{D}_{\mathrm{D}}$ containing additional patterns that the teacher can forecast reliably. Together, these subsets supplement the limited spatio-temporal dataset $\mathcal{D}_{\mathrm{ST}}$.

\noindent\textbf{Teacher-Student Setup.}
The pretrained TSFM serves as a frozen teacher, processing each node-wise and each auxiliary temporal sequence independently to provide intermediate temporal representations and forecasts. The STFM student shares its temporal modeling parameters across both sequence types. It incorporates spatial interactions when processing spatio-temporal samples and bypasses these interactions for temporal-only samples, allowing auxiliary temporal sequences to support temporal learning without introducing artificial spatial dependencies.


\noindent\textbf{Knowledge Distillation.}
We distill teacher knowledge via two complementary components:
\textit{Temporal Relation Distillation} and
\textit{Multi-Scale Prediction Distillation}.

\textit{Temporal Relation Distillation.}
This module transfers relationships among temporal patches across
selected teacher--student layers. By aligning pairwise similarities
derived from intermediate representations, it encourages the student
to preserve the teacher's temporal relational structure while
accommodating differences in hidden dimensions and patch counts.

\textit{Multi-Scale Prediction Distillation.}
This module aligns teacher and student forecasts at multiple temporal
resolutions through temporal aggregation. It complements temporal
relation distillation by transferring forecasting behavior at the
output level, capturing both fine temporal variations and broader trends.

The student is jointly trained on $\mathcal{D}_{\mathrm{ST}}$, $\mathcal{D}_{\mathrm{M}}$, and $\mathcal{D}_{\mathrm{D}}$ using ground-truth supervision and both distillation objectives.

\subsection{Temporal Data Selection}
\label{sec:temporal_selection}

As shown in Figure~\ref{fig2overview}(a), a shared pretrained encoder maps ST and TS windows into a common representation space. Temporal pattern discovery and exemplar selection establish an ST exemplar reference, and reference calibration determines the matching threshold. Auxiliary time series candidates are then scored by their proximity to this reference: nearby candidates undergo matched selection, while distant candidates pass through teacher reliability filtering, novel pattern discovery, and diversity selection. The selected embeddings are used to retrieve the corresponding temporal windows by index.

\subsubsection{Shared Temporal Representation}
\label{sec:selection_encoder}

We construct encoder pretraining samples from the limited spatio-temporal data $\mathcal{D}_{\mathrm{ST}}$ by treating each node's observations as an independent time series, omitting spatial relations. Let $\mathcal{D}'_{\mathrm{ST}}$ denote a fixed collection of these sampled node-wise windows.

We consider three alternative implementations of the pretrained encoder $E_{\psi}$ in Figure~\ref{fig2overview}(a): TS2Vec~\cite{yue2022ts2vec}, TS-TCC~\cite{ijcai2021-324}, and CoST~\cite{woo2022cost}. Each encoder is pretrained separately on the same $\mathcal{D}'_{\mathrm{ST}}$, with the preprocessing described below, using its own self-supervised objective and is then frozen throughout selection. Within each encoder configuration, the same frozen $E_{\psi}$ processes both ST and TS windows, placing samples from the two sources in a shared representation space.

For a training window $u=(\mathbf{x},\mathbf{y})$, where $\mathbf{x}\in\mathbb{R}^{L\times1}$ and $\mathbf{y}\in\mathbb{R}^{H\times1}$ are the history and observed target, we concatenate them into a trajectory $[\mathbf{x};\mathbf{y}]\in\mathbb{R}^{(L+H)\times1}$. Mean aggregation to a common sampling interval produces $\mathbf{s}(u)\in\mathbb{R}^{T\times1}$, where $T$ is the aligned sequence length. We then compute
\begin{align}
\widetilde{\mathbf{s}}(u)
&= \frac{\mathbf{s}(u)-\mu(\mathbf{s}(u))}{\sigma(\mathbf{s}(u))}
\label{eq:selection_embedding_s}\\
\mathbf{h}(u)
&= \operatorname{Pool}\left(E_{\psi}(\widetilde{\mathbf{s}}(u))\right)
\label{eq:selection_embedding_h}\\
\mathbf{z}(u)
&=\frac{\mathbf{h}(u)}{\lVert \mathbf{h}(u)\rVert_2}
\label{eq:selection_embedding_z}
\end{align}
where $\mu(\mathbf{s}(u))$ and $\sigma(\mathbf{s}(u))$ are the scalar mean and standard deviation of the aggregated trajectory. The encoder produces $E_{\psi}(\widetilde{\mathbf{s}}(u))\in\mathbb{R}^{T_{\psi}\times d}$, where $T_{\psi}$ is the encoder's output sequence length. The operator $\operatorname{Pool}$ performs global max pooling, yielding $\mathbf{h}(u)\in\mathbb{R}^{d}$. The final $L_2$ normalization produces $\mathbf{z}(u)$. These vectors $\mathbf{z}(u)$ form the ST Embeddings or TS Embeddings in Figure~\ref{fig2overview}(a) and are used for subsequent pattern discovery and reference proximity scoring.

\subsubsection{ST Exemplar Construction and Threshold Calibration}
\label{sec:reference_calibration}

We construct ST reference exemplars to represent temporal patterns in the limited ST data and calibrate a proximity threshold to support both matched and diversity selection. Following Figure~\ref{fig2overview}(a), this stage comprises temporal pattern discovery, exemplar selection, reference calibration, and reference proximity scoring.

\begin{figure}
    \centering
    \includegraphics[width=\linewidth]{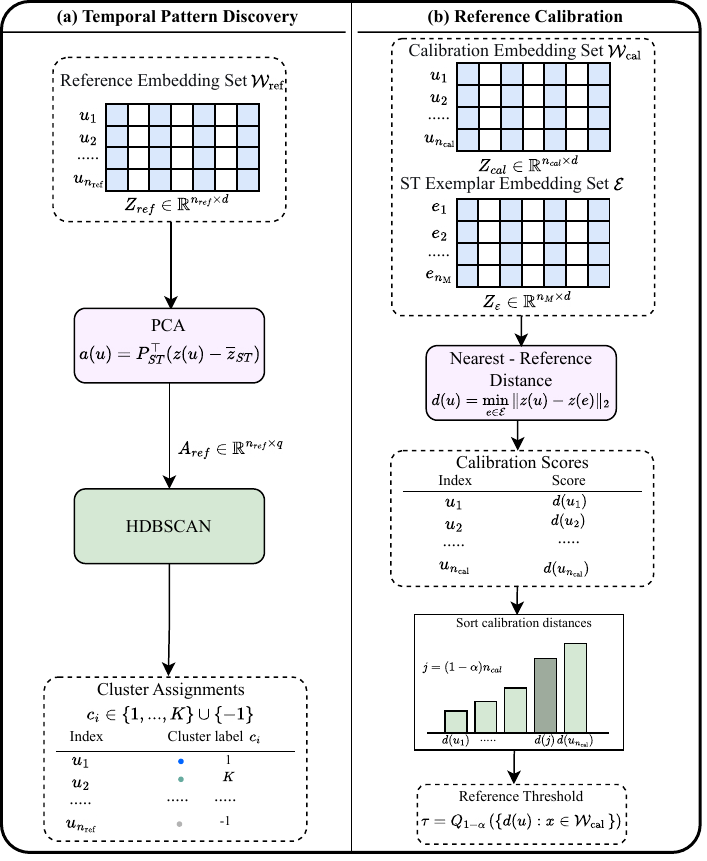}
    \caption{ST reference exemplar construction and threshold calibration.}
    \label{fig:pattern discovery}
    \vspace{-4mm}
\end{figure}

\textbf{Temporal pattern discovery.} As illustrated in Figure~\ref{fig:pattern discovery}(a), temporal pattern discovery maps ST reference embeddings to cluster assignments while preserving the identity of each window. Using a fixed random seed, we partition the available node-wise ST training windows $\mathcal{W}_{\mathrm{ST}}$ into disjoint reference and calibration sets, $\mathcal{W}_{\mathrm{ref}}$ and $\mathcal{W}_{\mathrm{cal}}$. The reference set is used to fit the clustering transformation, discover temporal patterns, and select exemplars; the calibration set is reserved for subsequent threshold estimation.

Let $\mathcal{W}_{\mathrm{ref}}=\{u_i\}_{i=1}^{n_{\mathrm{ref}}}$, where $n_{\mathrm{ref}}=|\mathcal{W}_{\mathrm{ref}}|$. Stacking the normalized embeddings $\mathbf{z}(u_i)^{\top}$ as rows gives $\mathbf{Z}_{\mathrm{ref}}\in\mathbb{R}^{n_{\mathrm{ref}}\times d}$. The row labels $u_i$ in Figure~\ref{fig:pattern discovery}(a) identify the original windows associated with these embeddings. We fit principal component analysis (PCA)~\cite{shlens2014tutorial} on $\mathbf{Z}_{\mathrm{ref}}$, with reference mean $\overline{\mathbf{z}}_{\mathrm{ST}}=n_{\mathrm{ref}}^{-1}\sum_{i=1}^{n_{\mathrm{ref}}}\mathbf{z}(u_i)\in\mathbb{R}^{d}$ and loading matrix $\mathbf{P}_{\mathrm{ST}}\in\mathbb{R}^{d\times q}$ containing the leading $q$ principal directions. The individual and stacked PCA coordinates are
\begin{equation}
\mathbf{a}(u)=\mathbf{P}_{\mathrm{ST}}^{\top}
\bigl(\mathbf{z}(u)-\overline{\mathbf{z}}_{\mathrm{ST}}\bigr)
\in\mathbb{R}^{q}
\label{eq}
\end{equation}
where $\overline{\mathbf{z}}_{\mathrm{ST}}\in\mathbb{R}^{d}$ is the mean reference embedding and $\mathbf{P}_{\mathrm{ST}}\in\mathbb{R}^{d\times q}$ contains the leading $q$ principal directions. This operation centers each embedding and projects its feature dimension from $d$ to $q$, with $q$ capped at the effective rank of the centered reference matrix. The resulting clustering coordinates form a matrix $\mathbf{A}_{\mathrm{ref}}$ in $\mathbb{R}^{n_{\mathrm{ref}}\times q}$.

We apply HDBSCAN~\cite{mcinnes2017accelerated,campello2013density} to the rows of $\mathbf{A}_{\mathrm{ref}}$ using Euclidean distance, fixed neighborhood and minimum-cluster-size parameters, and excess-of-mass cluster extraction. It determines the number of clusters $K$ and assigns each reference window a label
\begin{equation}
\mathcal{C}_k = \{u_i\in\mathcal{W}_{\mathrm{ref}} : c_i = k\},\quad k=-1,1,\ldots,K
\label{eq:selection_clusters}
\end{equation}
Here, $c_i=-1$ denotes density noise, as shown in the cluster assignments in Figure~\ref{fig:pattern discovery}(a), which is excluded from this selection. Each $\mathcal{C}_k$ contains the original windows assigned to one temporal pattern cluster and supplies candidates for exemplar selection. 

\textbf{Exemplar selection.}
Reference exemplars are constructed by random sampling stratified across the $K$ clusters. Selecting at least one exemplar from each of the $K$ clusters requires an exemplar budget $B_{\mathrm{E}}\geq K$. Using a fixed random seed, we first sample one window uniformly from each cluster. And then expand the exemplar set by drawing additional windows uniformly and without replacement from the clusters in turn, skipping exhausted clusters until the exemplar budget is reached or no candidates remain.


Let $\mathcal{E}_k\subseteq\mathcal{C}_k$ contain the exemplars selected from cluster $k$. The resulting ST reference exemplar set is
\begin{equation}
\begin{aligned}
\mathcal{E}
&= \bigcup_{k=1}^{K}\mathcal{E}_k\\
M
&= |\mathcal{E}|
= \min\left\{
B_{\mathrm{E}},
\sum_{k=1}^{K}|\mathcal{C}_k|
\right\}
\end{aligned}
\label{eqE}
\end{equation}

Each $e\in\mathcal{E}$ is a context--target window with normalized embedding $\mathbf{z}(e)\in\mathbb{R}^{d}$. Stacking these vectors gives the ST reference embeddings in Figure~\ref{fig2overview}(a), a matrix in $\mathbb{R}^{M\times d}$. The sampling procedure retains at least one exemplar from each cluster and distributes additional exemplars across the discovered patterns.

\textbf{Reference calibration.}
As illustrated in Figure~\ref{fig:pattern discovery}(b), reference calibration computes the distance from each calibration window to its nearest ST exemplar, sorts these distances, and determines the reference threshold. We define the proximity of a window $u$ to the ST reference as
\begin{equation}
 d(u)=\min_{e\in\mathcal E}\lVert \mathbf{z}(u)-\mathbf{z}(e)\rVert_2
\label{eq:selection_distance}
\end{equation}
Distances are computed using the normalized embeddings $\mathbf{z}$, while the PCA coordinates $\mathbf{a}$ are used for clustering. We evaluate $d(u)$ on the calibration windows and set
\begin{equation}
 \tau=Q_{1-\alpha}\!\left(\{d(u):u\in\mathcal{W}_{\mathrm{cal}}\}\right),
 \qquad 0<\alpha<1
\label{eq:selection_tau}
\end{equation}
where $n_{\mathrm{cal}}=|\mathcal{W}_{\mathrm{cal}}|$ and $Q_{1-\alpha}$ denotes the empirical $(1-\alpha)$-quantile, returning the $\lceil(1-\alpha)n_{\mathrm{cal}}\rceil$-th smallest calibration distance. Consequently, at least a fraction $1-\alpha$ of the calibration windows satisfy $d(u)\leq\tau$.

\textbf{Reference proximity scoring.}
With $\mathcal{E}$ and $\tau$ fixed, we compute $d(u)$ for every auxiliary candidate $u\in\mathcal{D}_{\mathrm{TS}}$. These distances correspond to the scores attached to TS embeddings in Figure~\ref{fig2overview}(a). Together with $\tau$, they guide matched selection within the reference neighborhood and distant-candidate screening for diversity selection. Both branches therefore use the same ST reference and proximity measure.

\subsubsection{Matched Data Selection}
\label{sec:matched_selection}
Matched data selection retrieves auxiliary windows close to the ST reference. Using the calibrated threshold, we define the matched pool as
\begin{equation}
 \mathcal U_{\mathrm{M}}=\{u\in\mathcal{D}_{\mathrm{TS}}:d(u)\leq\tau\}.
\label{eq:matched_pool}
\end{equation}
Each candidate is assigned to the cluster of its closest exemplar. Within each resulting bucket, candidates are ranked by increasing $d(u)$. We cycle through nonempty buckets in a fixed seeded order, taking the closest remaining eligible candidate from each, until the budget $B_{\mathrm{M}}$ is reached. The selected windows form the matched subset $\mathcal{D}_{\mathrm{M}}\subseteq\mathcal{U}_{\mathrm{M}}$. This limits domination by a single reference pattern.


\subsubsection{Diverse Data Selection}
\label{sec:diversity_selection}

Following Figure~\ref{fig2overview}(a), this branch comprises teacher reliability filtering, novel pattern discovery, and diversity selection. Its input pool contains auxiliary windows outside the expanded ST reference neighborhood:
\begin{equation}
 \mathcal U_{\mathrm{D}}^{0}=\{u\in\mathcal{D}_{\mathrm{TS}}:
 d(u)>\tau+\delta\},\qquad \delta>0
\label{eq:distant_pool}
\end{equation}
after excluding windows overlapping $\mathcal{D}_{\mathrm{M}}$. The margin $\delta$ separates the two branches; windows with $\tau<d(u)\leq\tau+\delta$ are used by neither.

\textbf{Teacher reliability filtering.}
Distance from the ST reference alone does not ensure reliable teacher supervision. For each $u=(\mathbf{x},\mathbf{y})\in\mathcal U_{\mathrm{D}}^{0}$, the frozen teacher predicts $\widehat{\mathbf{y}}^T=g_{\phi}(\mathbf{x})$, and we measure its forecasting error:
\begin{equation}
 \ell_T(u)=
 \frac{H^{-1}\sum_{h=1}^{H}|\widehat y_h^T-y_h|}
 {\max\{\sigma(\mathbf{x}),\epsilon_s\}}
\label{eq:teacher_reliability}
\end{equation}
where $\widehat y_h^T$ and $y_h$ denote the $h$-th scalar entries of $\widehat{\mathbf{y}}^T$ and $\mathbf{y}$, respectively. Here, $\sigma(\mathbf{x})$ is the history standard deviation and $\epsilon_s>0$ is a scale floor for source $s$. As in reference calibration, we set an acceptance threshold using the ST calibration windows:
\begin{equation}
 \begin{aligned}
 \eta &= Q_{1-\beta}\!\left(\{\ell_T(u):u\in\mathcal{W}_{\mathrm{cal}}\}\right)\\
 \mathcal U_{\mathrm{D}}^{1}
 &=\{u\in\mathcal U_{\mathrm{D}}^{0}:\ell_T(u)\leq\eta\}
 \end{aligned}
\label{eq:reliable_pool}
\end{equation}
Here, $0<\beta<1$. This retains candidates whose teacher error lies within the calibrated tolerance.

\textbf{Novel pattern discovery.}
After teacher reliability filtering, we fit a separate PCA transformation, again capped at the effective rank, on $\{\mathbf{z}(u):u\in\mathcal U_{\mathrm{D}}^{1}\}$ and apply HDBSCAN with specified minimum cluster size and neighborhood parameters to obtain $\{\mathcal G_j\}_{j=1}^{J}$. Refitting PCA here allows clustering to retain variation specific to the distant pool. We exclude density-noise points and retain members with strength at least $p_0$. Consequently, novelty means a coherent candidate pattern outside the calibrated ST reference neighborhood.

\textbf{Diversity selection.}
As in matched selection, we visit nonempty cluster pools in a fixed seeded round-robin order and apply the same overlap exclusion and tie-breaking rules. On a cluster's first visit, we select its eligible medoid, minimizing the sum of distances to the other eligible members. On subsequent visits, we choose the eligible window whose minimum distance to the already selected windows in that cluster is largest. All distances are Euclidean distances between the normalized embeddings $\mathbf{z}$. Selection stops at budget $B_{\mathrm{D}}$ or candidate exhaustion. The selected windows form the diverse subset $\mathcal{D}_{\mathrm{D}}\subseteq\mathcal U_{\mathrm{D}}^{1}$, with $|\mathcal{D}_{\mathrm{D}}|\leq B_{\mathrm{D}}$. Together, $\mathcal{D}_{\mathrm{M}}$ and $\mathcal{D}_{\mathrm{D}}$ supplement $\mathcal{D}_{\mathrm{ST}}$ during joint training.
All selection operations are offline and are summarized in Algorithm~\ref{alg:t2st_selection}.

\begin{algorithm}[!t]
\caption{Offline Temporal Data Selection in T2ST}
\label{alg:t2st_selection}
\small
\begin{algorithmic}[1]
\Require Training pools $\mathcal D_{\mathrm{ST}},\mathcal D_{\mathrm{TS}}$;
encoder $E_\psi$; frozen teacher $g_\phi$;
budgets $B_{\mathrm E},B_{\mathrm M},B_{\mathrm D}$; selection settings
\Ensure Matched subset $\mathcal D_{\mathrm M}$ and diverse subset $\mathcal D_{\mathrm D}$
\Statex Records are identified by dataset, series row, and window start.
Use a fixed seed for splits, ties, and round-robin cluster visits.

\State Construct $\mathcal W_{\mathrm{ST}}$; pretrain $E_\psi$ on $\mathcal D'_{\mathrm{ST}}$ and freeze it
\State Embed $\mathcal W_{\mathrm{ST}}\cup\mathcal D_{\mathrm{TS}}$ using \eqref{eq:selection_embedding_s}--\eqref{eq:selection_embedding_z}
\State Split $\mathcal W_{\mathrm{ST}}$ into $\mathcal W_{\mathrm{ref}}$ and $\mathcal W_{\mathrm{cal}}$
\State Fit PCA--HDBSCAN to build $\mathcal E$ by \eqref{eqE}
\State Compute $d,\tau$ by \eqref{eq:selection_distance}--\eqref{eq:selection_tau}; initialize $\mathcal D_{\mathrm M},\mathcal D_{\mathrm D}\gets\varnothing$
\State Form $\mathcal U_{\mathrm M}$ by \eqref{eq:matched_pool}; group by nearest-exemplar cluster into buckets $\mathcal B_k$, sorted by increasing $d(u)$
\While{$|\mathcal D_{\mathrm M}|<B_{\mathrm M}$ and $\bigcup_k\mathcal B_k\neq\varnothing$}
    \State Visit the next $\mathcal B_k$; add its closest remaining record $u$ to $\mathcal D_{\mathrm M}$
    \State Remove the selected record identity from all $\mathcal B_i$
\EndWhile
\State Form $\mathcal U_{\mathrm D}^{0}$ by \eqref{eq:distant_pool}, excluding identities already in $\mathcal D_{\mathrm M}$
\State Compute $\ell_T$, calibrate $\eta$, and form $\mathcal U_{\mathrm D}^{1}$ by \eqref{eq:teacher_reliability}--\eqref{eq:reliable_pool}
\State \textbf{if} $\mathcal U_{\mathrm D}^{1}$ cannot support clustering, \textbf{return} $(\mathcal D_{\mathrm M},\varnothing)$
\State Refit PCA--HDBSCAN on the reliable embeddings to obtain nonnoise clusters $\mathcal G_j$
\State Retain members with strength $\geq p_0$ in $\mathcal Q_j\subseteq\mathcal G_j$;
\While{$|\mathcal D_{\mathrm D}|<B_{\mathrm D}$ and $\bigcup_j\mathcal Q_j\neq\varnothing$}
    \State Visit the next nonempty $\mathcal Q_j$; choose $u$ as its medoid if $\mathcal S_j=\varnothing$, otherwise maximize its minimum distance to $\mathcal S_j$
    \State Add $u$ to $\mathcal S_j$ and $\mathcal D_{\mathrm D}$; remove its record identity from all $\mathcal Q_i$
\EndWhile
\State \Return $\mathcal D_{\mathrm M},\mathcal D_{\mathrm D}$
\end{algorithmic}
\end{algorithm}

\subsection{Knowledge Distillation}
\label{sec:knowledge_distillation}

After temporal data selection, T2ST trains an STFM student using the limited spatio-temporal dataset $\mathcal{D}_{\mathrm{ST}}$ and the selected auxiliary subsets $\mathcal{D}_{\mathrm{M}}$ and $\mathcal{D}_{\mathrm{D}}$. A frozen TSFM provides two complementary forms of supervision: relationships among historical temporal patches and predictions at multiple temporal resolutions. The former transfers the teacher's organization of temporal context, while the latter transfers its forecasting behavior. We first describe the teacher and the student, followed by the knowledge transfer objectives.

\subsubsection{Frozen TSFM Teacher}
\label{sec:kd_teacher}

Let $g_{\phi}$ denote a pretrained TSFM with fixed parameters $\phi$. We describe the scalar forecasting setting used in our implementation, with $\mathbf{X}\in\mathbb{R}^{L\times N\times 1}$ and $\mathbf{Y}\in\mathbb{R}^{H\times N\times 1}$. For a spatio-temporal sample, the teacher processes each node history $\mathbf{x}_n=\mathbf{X}_{:,n,1}$ independently:
\begin{equation}
 \widehat{\mathbf{y}}^{T}_{n}=g_{\phi}(\mathbf{x}_n),
 \qquad n=1,\ldots,N
 \label{eq:kd_teacher_prediction}
\end{equation}
Stacking these outputs gives $\widehat{\mathbf{Y}}^{T}\in\mathbb{R}^{H\times N\times 1}$. An auxiliary temporal window is processed in the same way with $N=1$. Thus, the teacher supplies temporal supervision for both data types without requiring spatial structure.

In addition to forecasts, we extract hidden representations $\mathbf{H}^{T}_{\ell,n}\in\mathbb{R}^{P_T\times d_T}$ from selected teacher layers, where $P_T$ is the number of history patches and $d_T$ is the hidden dimension. We instantiate the teacher from a local copy of the official Google Research TimesFM-2.5 checkpoint and extract representations from three layers. We perform no additional teacher pretraining or fine-tuning. All teacher parameters remain frozen during offline reliability filtering and student distillation.

\subsubsection{STFM Student with Shared Temporal Modeling}
\label{sec:kd_student}

The student $f_{\theta}$ retains the temporal and spatial modeling components of its STFM backbone. It supports two input-dependent computation paths:
\begin{equation}
 \widehat{\mathbf{Y}}^{S}=
 \begin{cases}
 f_{\theta}^{\mathrm{ST}}(\mathbf{X},\mathcal{G}),
     & (\mathbf{X},\mathbf{Y})\in\mathcal{D}_{\mathrm{ST}}\\
 f_{\theta}^{\mathrm{T}}(\mathbf{X}),
     & (\mathbf{X},\mathbf{Y})\in\mathcal{D}_{\mathrm{M}}\cup\mathcal{D}_{\mathrm{D}}
 \end{cases}
 \label{eq:kd_student_paths}
\end{equation}
Here, $f_{\theta}^{\mathrm{ST}}$ and $f_{\theta}^{\mathrm{T}}$ denote two execution paths of the same student, sharing temporal modeling and forecasting parameters. The ST path uses the available spatial structure $\mathcal{G}$. The temporal-only path bypasses cross-node interactions and omits graph-dependent spatial embeddings. Auxiliary training therefore updates the shared temporal pathway, while ST samples provide supervision for spatial modeling.

For relation distillation, the student exposes temporal representations $\mathbf{H}^{S}_{i,n}\in\mathbb{R}^{P_S\times d_S}$ at selected layers. In the OpenCity instantiation, these features are taken after temporal modeling and before graph convolution within each block. On ST inputs, they can retain spatial information from context embeddings and preceding blocks. The formulation permits another STFM backbone, such as UrbanFM, provided that its adapter exposes temporally ordered historical features, enables temporal-only processing with shared parameters, and produces forecasts for the same target horizon.
\subsubsection{Knowledge Transfer Objectives}
\label{sec:kd_objectives}
We transfer temporal knowledge from the frozen TSFM teacher to the STFM student through two complementary objectives: temporal relation distillation and multi-scale prediction distillation. The former preserves relationships among intermediate temporal patches, while the latter aligns teacher and student forecasts at multiple temporal resolutions, providing supervision at both the representation and prediction levels.

\noindent\textbf{Temporal relation distillation.}
Directly matching hidden features is restrictive because the teacher and student can differ in both temporal resolution and representation dimension. Instead, we align pairwise patch similarities separately for each sample and node. Let $\mathcal{P}$ be the set of selected student--teacher layer pairs $(i,\ell)$. Per pair, adaptive average pooling (AAP) aligns the student's patch axis to the teacher's patch count:
\begin{equation}
 \widetilde{\mathbf{H}}^{S}_{i,n}
 =\operatorname{AAP}_{P_T}(\mathbf{H}^{S}_{i,n})
 \in\mathbb{R}^{P_T\times d_S}
 \label{eq:kd_patch_alignment}
\end{equation}
The operator is the identity when $P_S=P_T$. It acts only along the temporal patch axis, leaving the student's feature dimension unchanged.

Let $\mathcal{N}(\mathbf{H})$ normalize each row $\mathbf{h}$ as $\mathbf{h}/\max(\lVert\mathbf{h}\rVert_2,\epsilon)$, with $\epsilon>0$. The temporal relation matrices are
\begin{equation}
 \begin{aligned}
 \mathbf{R}^{S}_{i,n}
 &=\mathcal{N}(\widetilde{\mathbf{H}}^{S}_{i,n})
   \mathcal{N}(\widetilde{\mathbf{H}}^{S}_{i,n})^{\top}\\
 \mathbf{R}^{T}_{\ell,n}
 &=\mathcal{N}(\mathbf{H}^{T}_{\ell,n})
   \mathcal{N}(\mathbf{H}^{T}_{\ell,n})^{\top}
 \end{aligned}
 \label{eq:kd_relations}
\end{equation}
Both matrices belong to $\mathbb{R}^{P_T\times P_T}$, and each entry measures the similarity of two temporal patches within the same node history. For a mini-batch of $B$ samples, the relation loss is
\begin{equation}
 \mathcal{L}_{\mathrm{rel}}
 =\frac{1}{|\mathcal{P}|BNP_T^2}
 \sum_{(i,\ell)\in\mathcal{P}}\sum_{b=1}^{B}\sum_{n=1}^{N}
 \left\lVert \mathbf{R}^{S}_{i,b,n}
       -\mathbf{R}^{T}_{\ell,b,n}\right\rVert_F^2
 \label{eq:kd_relation_loss}
\end{equation}
For the three-block OpenCity configuration, we use $\mathcal{P}=\{(1,6),(2,13),(3,20)\}$.

\noindent\textbf{Multi-scale prediction distillation.}
Relation alignment constrains historical representations but does not directly constrain future forecasts. We therefore align teacher and student predictions at multiple temporal resolutions. 

For an aggregation factor $r$, define nonoverlapping temporal averaging as
\begin{equation}
 [\mathcal{A}_{r}(\widehat{\mathbf{Y}})]_{b,k,n,1}
 =\frac{1}{r}\sum_{j=1}^{r}
   [\widehat{\mathbf{Y}}]_{b,(k-1)r+j,n,1}
 \label{eq:kd_temporal_aggregation}
\end{equation}
where $k=1,\ldots,H/r$. We use $\mathcal{R}=\{1,2,4,8\}$ and forecasting horizons divisible by every factor. The loss is
\begin{equation}
 \mathcal{L}_{\mathrm{ms}}
 =\frac{1}{|\mathcal{R}|}\sum_{r\in\mathcal{R}}
 \operatorname{MAE}\!\left(
 \mathcal{A}_{r}(\widehat{\mathbf{Y}}^{S}),
 \mathcal{A}_{r}(\widehat{\mathbf{Y}}^{T})\right)
 \label{eq:kd_multiscale_loss}
\end{equation}
The $r=1$ term transfers step-level forecasting behavior, while coarser averages constrain broader temporal variations. Each resolution contributes equally after averaging its prediction errors.

\noindent\textbf{Joint training and inference.}
Both distillation losses are applied to ST, matched, and diversity batches. For a batch $\mathcal{B}$, we combine them with ground-truth forecasting supervision:
\begin{equation}
 \mathcal{L}(\mathcal{B})
 =\mathcal{L}_{\mathrm{sup}}(\mathcal{B})
 +\lambda_{\mathrm{rel}}\mathcal{L}_{\mathrm{rel}}(\mathcal{B})
 +\lambda_{\mathrm{ms}}\mathcal{L}_{\mathrm{ms}}(\mathcal{B})
 \label{eq:kd_batch_objective}
\end{equation}
where $\mathcal{L}_{\mathrm{sup}}$ is the student's supervised forecasting loss and $\lambda_{\mathrm{rel}},\lambda_{\mathrm{ms}}$ balance the two transfer objectives.

Mixed training is organized around ST batches. Let $I_{\mathrm{M},t}$ and $I_{\mathrm{D},t}$ indicate whether matched and diverse data are scheduled at training event $t$, respectively, with $I_{\mathrm{D},t}\leq I_{\mathrm{M},t}$. With combined updates, the event objective is
\begin{equation}
 \begin{aligned}
 \mathcal{J}_{t}
 ={}&\mathcal{L}(\mathcal{B}_{\mathrm{ST},t})\\
 &+I_{\mathrm{M},t}\gamma_{\mathrm{M}}
   \sum_{j=1}^{K_{\mathrm{M}}}
   \mathcal{L}(\mathcal{B}_{\mathrm{M},t,j})\\
 &+I_{\mathrm{D},t}\widetilde{\gamma}_{\mathrm{D},t}
   \mathcal{L}(\mathcal{B}_{\mathrm{D},t})
 \end{aligned}
 \label{eq:kd_joint_objective}
\end{equation}
Here, $K_{\mathrm{M}}$ is the number of matched batches per scheduled event and $\gamma_{\mathrm{M}}$ is their per-batch weight. To control the diversity contribution, let $A_t=\gamma_{\mathrm{M}}\sum_j\mathcal{L}(\mathcal{B}_{\mathrm{M},t,j})$ and $D_t=\mathcal{L}(\mathcal{B}_{\mathrm{D},t})$. On diversity events, its effective weight is
\begin{equation}
 \widetilde{\gamma}_{\mathrm{D},t}
 =\gamma_{\mathrm{D}}\min\!\left\{
 1,\frac{\rho\,A_t}
 {\gamma_{\mathrm{D}}D_t+\epsilon}
 \right\}
 \label{eq:kd_diversity_weight}
\end{equation}
where $0<\rho<1$ caps the weighted diversity loss relative to the weighted matched loss, and $0<\gamma_{\mathrm{D}}<\gamma_{\mathrm{M}}$. This coefficient is treated as constant during backpropagation. Sampling intervals and loss weights therefore control auxiliary exposure and its contribution to training. Ground-truth supervision anchors the student to observed outcomes while the frozen teacher supplies additional temporal guidance. 

\section{Experiment}
We evaluate forecasting with limited structured observations through five research questions:
\begin{itemize}
    \item \textbf{RQ1 (Overall Effectiveness):}  How does T2ST compare with its STFM backbones trained on the same limited ST subset or the full ST training set?
    \item \textbf{RQ2 (Temporal Data Selection):} How do the composition of auxiliary temporal data and the choice of selection encoder affect T2ST’s forecasting performance?
    \item \textbf{RQ3 (Batch Mixing Ratio):} How does the ratio of ST to matched temporal batches affect forecasting?
    \item \textbf{RQ4 (ST Data Budget):}  How does T2ST’s forecasting performance vary across different ST data budgets?
    \item \textbf{RQ5 (Distillation Weight Sensitivity):} How sensitive is T2ST to the loss weights?
\end{itemize}

\subsection{Experiment Setup}
\label{sec:experiment_setup}

\subsubsection{Evaluation Settings and Data Budgets}
We consider two evaluation settings, summarized in Table~\ref{tab:evaluation-datasets}. The OpenCity setting~\cite{li2025open} evaluates long-term forecasting on CAD3, CAD5, PEMS07M, TrafficSH, CHI-TAXI, and NYC-BIKE. It uses one day of observations to predict the next day, corresponding to 288 steps for five-minute data. The UrbanFM setting~\cite{chen2026urbanfm} uses a 12-step history and a 12-step horizon on the five-minute data packages of EvalST, whose 12 datasets cover flow, occupancy, speed, and grid-based urban observations. 
\begin{table}[!t]
\centering
\caption{Evaluation datasets for short-term forecasting and long-term forecasting.}
\label{tab:evaluation-datasets}
\begingroup
\footnotesize
\setlength{\tabcolsep}{2pt}
\renewcommand{\arraystretch}{1.0}
\resizebox{\columnwidth}{!}{%
\begin{tabular}{@{}lllrc@{}}
\toprule
\textbf{Dataset} & \textbf{Category} & \textbf{Data format} & \textbf{\# Regions} & \textbf{Interval} \\
\midrule
\multicolumn{5}{@{}l}{\textbf{(a) Short-term: $L=12$, $H=12$}} \\
\midrule
PEMS-03 & Flow & Sensor-based & 358 & 5 min \\
PEMS-04 & Flow & Sensor-based & 307 & 5 min \\
PEMS-07 & Flow & Sensor-based & 883 & 5 min \\
PEMS-08 & Flow & Sensor-based & 170 & 5 min \\
OCC-Pairs & Occupancy & Sensor-based & 106 & 5 min \\
OCC-Hamburg & Occupancy & Sensor-based & 240 & 5 min \\
PEMS-BAY & Speed & Sensor-based & 325 & 5 min \\
METR-LA & Speed & Sensor-based & 207 & 5 min \\
Traffic-SH & Road & Grid-based & 896 & 5 min \\
Bike-NYC & Bike & Grid-based & 200 & 5 min \\
Taxi-NYC & Taxi & Grid-based & 200 & 5 min \\
Tdrive-BJ & Trajectory & Grid-based & 896 & 5 min \\
\midrule
\multicolumn{5}{@{}l}{\textbf{(b) Long-term: one-day history and horizon}} \\
\midrule
CAD3 & Traffic flow & Sensor-based & 480 & 5 min \\
CAD5 & Traffic flow & Sensor-based & 211 & 5 min \\
CHI-TAXI & Taxi demand & Grid-based & 77 & 30 min \\
NYC-BIKE & Bike trajectory & Grid-based & 540 & 30 min \\
TrafficSH & Traffic speed & Grid-based & 896 & 30 min \\
PEMS07M & Traffic speed & Sensor-based & 228 & 5 min \\
\bottomrule
\end{tabular}%
}
\endgroup
\vspace{-4mm}
\end{table}

\begin{table*}[!t]
    \centering
    \caption{\textbf{Long-term forecasting} results. The first ten methods
    are full-shot baselines.Best and second-best results are shown in bold and underlined, respectively, with light purple cell backgrounds.}
    \label{tab:full-shot-comparison}
    \begingroup
    \definecolor{T2STLightPurple}{HTML}{EFE7F7}
    \setlength{\tabcolsep}{2.5pt}
    \renewcommand{\arraystretch}{1.15}
    \resizebox{\textwidth}{!}{%
    \begin{tabular}{l*{12}{c}}
        \toprule
        \multirow{2}{*}{\textbf{Model}}
        & \multicolumn{2}{c}{\textbf{CAD3}}
        & \multicolumn{2}{c}{\textbf{CAD5}}
        & \multicolumn{2}{c}{\textbf{PEMS07M}}
        & \multicolumn{2}{c}{\textbf{TrafficSH}}
        & \multicolumn{2}{c}{\textbf{CHI-TAXI}}
        & \multicolumn{2}{c}{\textbf{NYC-BIKE}} \\
        \cmidrule(lr){2-3}
        \cmidrule(lr){4-5}
        \cmidrule(lr){6-7}
        \cmidrule(lr){8-9}
        \cmidrule(lr){10-11}
        \cmidrule(lr){12-13}
        & \textbf{\textit{MAE}} & \textbf{\textit{RMSE}}
        & \textbf{\textit{MAE}} & \textbf{\textit{RMSE}}
        & \textbf{\textit{MAE}} & \textbf{\textit{RMSE}}
        & \textbf{\textit{MAE}} & \textbf{\textit{RMSE}}
        & \textbf{\textit{MAE}} & \textbf{\textit{RMSE}}
        & \textbf{\textit{MAE}} & \textbf{\textit{RMSE}} \\
        \midrule
        PDFormer
        & 20.28 & 36.43 & 12.89 & 21.18
        & 4.62 & 8.36 & 0.77 & 1.23
        & 4.03 & 12.82 & 7.33 & 13.01 \\
        STWA
        & 21.65 & 37.55 & 14.43 & 24.14
        & 4.54 & 8.57 & 1.42 & 2.49
        & 3.70 & 11.49 & 7.97 & 14.35 \\
        MSDR
        & 22.74 & 37.15 & 15.44 & 29.52
        & 4.81 & 8.61 & 1.84 & 3.55
        & 3.55 & 10.39 & 8.08 & 13.97 \\
        AGCRN
        & 18.72 & 31.93 & 12.88 & 23.78
        & 4.61 & 8.63 & 1.33 & 2.45
        & 3.60 & 11.31 & 8.13 & 14.46 \\
        MTGNN
        & 17.59 & 28.92 & 11.70 & 20.30
        & 4.52 & 8.06 & 0.81 & 1.26
        & 3.27 & 9.87 & 6.48 & 11.49 \\
        STSGCN
        & 21.88 & 34.52 & 13.87 & 22.32
        & 4.56 & 8.05 & 1.66 & 3.33
        & 4.87 & 14.40 & 6.85 & 11.98 \\
        GWN
        & 16.94 & 28.81
        & \cellcolor{T2STLightPurple}\textbf{10.69} & 19.75
        & \cellcolor{T2STLightPurple}\underline{4.17}
        & \cellcolor{T2STLightPurple}\underline{7.84}
        & 0.76 & 1.39
        & 3.56 & 11.27
        & 7.61 & 13.56 \\
        ASTGNN
        & 23.60 & 39.35 & 12.58 & 21.23
        & 4.39 & 8.21 & 0.69 & 1.09
        & 3.28 & 10.32 & 6.44 & 11.62 \\
        STGCN
        & 20.24 & 34.34 & 13.76 & 27.49
        & 4.44 & 8.37 & 1.60 & 3.14
        & 3.09 & 9.54 & 8.01 & 13.94 \\
        TGCN
        & 19.56 & 30.82 & 13.07 & 21.56
        & 4.88 & 8.38 & 1.79 & 2.65
        & 4.02 & 11.70 & 7.75 & 13.49 \\
        \hline
        UrbanFM(10\%)
        & 20.86 & 36.53 & 13.48 & 22.54
        & 5.48 & 10.50 & 0.66 & 0.93
        & 2.68 & 6.35 & 7.68 & 14.53 \\
        UrbanFM
        & 20.25 & 35.88 & 12.29 & 21.07
        & 5.13 & 9.93
        & \cellcolor{T2STLightPurple}\textbf{0.55}
        & \cellcolor{T2STLightPurple}\textbf{0.84}
        & 2.85 & 7.85 & 7.33 & 13.93 \\
        \textbf{T2ST (UrbanFM)}
        & 19.64 & 34.54 & 12.30 & 20.99
        & 4.88 & 9.31
        & \cellcolor{T2STLightPurple}\textbf{0.55} & 0.88
        & 2.78 & 7.47 & 7.51 & 14.28 \\
        \hline
        OpenCity (10\%)
        & 17.97 & 31.62 & 11.73 & 20.33
        & 4.59 & 8.32
        & \cellcolor{T2STLightPurple}\underline{0.58} & 0.89
        & \cellcolor{T2STLightPurple}\underline{2.29} & 5.71
        & 6.20 & 11.24 \\
        OpenCity
        & \cellcolor{T2STLightPurple}\underline{16.53}
        & \cellcolor{T2STLightPurple}\underline{28.34}
        & 11.30
        & \cellcolor{T2STLightPurple}\underline{19.42}
        & 4.53 & 8.30
        & \cellcolor{T2STLightPurple}\textbf{0.55}
        & \cellcolor{T2STLightPurple}\underline{0.85}
        & \cellcolor{T2STLightPurple}\textbf{1.99}
        & \cellcolor{T2STLightPurple}\textbf{4.64}
        & \cellcolor{T2STLightPurple}\underline{6.13}
        & \cellcolor{T2STLightPurple}\underline{11.22} \\
        \textbf{T2ST (OpenCity)}
        & \cellcolor{T2STLightPurple}\textbf{16.25}
        & \cellcolor{T2STLightPurple}\textbf{27.72}
        & \cellcolor{T2STLightPurple}\underline{10.74}
        & \cellcolor{T2STLightPurple}\textbf{18.51}
        & \cellcolor{T2STLightPurple}\textbf{4.16}
        & \cellcolor{T2STLightPurple}\textbf{7.81}
        & \cellcolor{T2STLightPurple}\textbf{0.55}
        & \cellcolor{T2STLightPurple}\underline{0.85}
        & 2.34
        & \cellcolor{T2STLightPurple}\underline{5.41}
        & \cellcolor{T2STLightPurple}\textbf{6.05}
        & \cellcolor{T2STLightPurple}\textbf{10.92} \\
        \bottomrule
    \end{tabular}%
    }
    \endgroup
\end{table*}

\definecolor{T2STLightPurple}{HTML}{EFE7F7}

\newcommand{\bestval}[1]{%
  \cellcolor{T2STLightPurple}\textbf{#1}%
}
\newcommand{\secondval}[1]{%
  \cellcolor{T2STLightPurple}\underline{#1}%
}
\begin{table*}[!t]
\centering
\caption{\textbf{Short-term forecasting} results.
Best and second-best results are shown in bold and underlined,
respectively, with light purple cell backgrounds.}
\label{tab:urbanfm-10pct-5min}
\begingroup
\setlength{\tabcolsep}{4pt}
\renewcommand{\arraystretch}{1.15}
\resizebox{\textwidth}{!}{%
\begin{tabular}{l*{12}{c}}
\toprule
\multirow{2}{*}{\textbf{Model}}
& \multicolumn{3}{c}{\textbf{TrafficSH}}
& \multicolumn{3}{c}{\textbf{NYC-BIKE}}
& \multicolumn{3}{c}{\textbf{NYC-TAXI}}
& \multicolumn{3}{c}{\textbf{Tdrive-BJ}} \\
\cmidrule(lr){2-4}
\cmidrule(lr){5-7}
\cmidrule(lr){8-10}
\cmidrule(lr){11-13}
& \textit{MAE} & \textit{RMSE} & \textit{MAPE}
& \textit{MAE} & \textit{RMSE} & \textit{MAPE}
& \textit{MAE} & \textit{RMSE} & \textit{MAPE}
& \textit{MAE} & \textit{RMSE} & \textit{MAPE} \\
\midrule
TimesFM++
& 0.49 & 1.14 & 7.31
& 3.98 & 7.37 & 65.52
& 8.95 & 20.04 & 49.50
& 13.90 & 30.48 & 18.26 \\
Moirai
& 0.46 & 0.91 & 8.76
& 1.63 & 5.87 & 85.18
& 7.58 & 22.72 & 63.33
& 21.73 & 52.16 & 31.99 \\
TimeMoE
& 0.43 & \secondval{0.74} & \secondval{6.65}
& 3.90 & 8.08 & 64.20
& 10.05 & 23.59 & 89.85
& 21.40 & 36.67 & 30.76 \\
Chronos
& 0.45 & 0.80 & 6.77
& 4.08 & 7.64 & 57.62
& 9.48 & 21.69 & \secondval{43.22}
& 16.91 & 35.01 & 20.79 \\
FactoST
& \bestval{0.40} & \bestval{0.70} & \bestval{5.84}
& 3.77 & 7.11 & \bestval{54.36}
& 8.84 & 21.68 & \bestval{42.33}
& 17.26 & 35.23 & 20.61 \\
\hline
OpenCity (10\%)
& 0.43 & 0.93 & 7.57
& 1.26 & 4.50 & 62.15
& 5.49 & 17.32 & 50.07
& 12.65 & 37.57 & 18.36 \\
OpenCity
& 0.44 & 0.97 & 7.65
& 1.29 & 4.67 & 65.25
& 5.45 & 17.04 & 52.01
& 11.99 & 33.87 & \bestval{17.69} \\
\textbf{T2ST (OpenCity)}
& \secondval{0.41} & 0.88 & 7.11
& 1.23 & 4.43 & 60.69
& \bestval{5.28} & \bestval{16.77} & 47.37
& 12.25 & 35.49 & \secondval{17.92} \\
\hline
UrbanFM (10\%)
& 0.50 & 0.95 & 7.79
& \bestval{1.19} & \bestval{4.11} & \secondval{56.45}
& 5.51 & 17.62 & 54.63
& \secondval{11.37} & \secondval{29.16} & 25.18 \\
UrbanFM
& 0.47 & 0.90 & 7.54
& \secondval{1.20} & \bestval{4.11} & 57.29
& 5.43 & 17.33 & 55.36
& 12.38 & 30.26 & 33.92 \\
\textbf{T2ST (UrbanFM)}
& 0.47 & 0.94 & 7.67
& \bestval{1.19} & \secondval{4.24} & 58.95
& \secondval{5.34} & \secondval{16.89} & 50.78
& \bestval{10.53} & \bestval{28.86} & 19.09 \\
\bottomrule
\end{tabular}%
}
\endgroup
\end{table*}

In pre-training, long-term ST data follow OpenCity, and short-term ST data follow UrbanFM's WorldST collection. The 10\% ST budget is measured against eligible samples on the complete source timelines, with all selected samples drawn from the training splits. T2ST supplements this restricted subset $\mathcal{D}_{\mathrm{ST}}$ with matched and diverse temporal windows from the GiftEvalPretrain corpus~\cite{aksu2024giftevalbenchmarkgeneraltime}.

\subsubsection{Baselines and Model Configurations}
In both settings, we apply T2ST as a training framework to two STFM backbones, OpenCity and UrbanFM. For each backbone, we compare the T2ST-trained model with its counterparts trained solely on the restricted or the full ST data. Long-term baselines also include ten full-shot task-specific predictors, such as PDFormer~\cite{jiang2023pdformer}, AGCRN~\cite{bai2020adaptive}, MTGNN~\cite{wu2020connecting}, and GWN~\cite{wu2019graph}. Short-term baselines include TimesFM++, Moirai, TimeMoE~\cite{shi2025time}, Chronos, and FactoST~\cite{zhong2026learning}.

T2ST uses a frozen TimesFM~2.5 teacher and TS2Vec~\cite{yue2022ts2vec} for offline temporal-data selection. Before distillation, the STFM student is initialized from a checkpoint trained on the restricted 10\% ST subset. The student is then distilled using temporal-relation and multi-scale prediction losses. The encoder study replaces TS2Vec with TS-TCC~\cite{ijcai2021-324} or CoST~\cite{woo2022cost}.

\subsubsection{Evaluation Metrics}
We report mean absolute error (MAE) and root mean squared error (RMSE) for all datasets. For short-term forecasting on grid-based datasets, we additionally report mean absolute percentage error (MAPE).
\subsection{Overall Forecasting Performance (RQ1)}
\label{sec:experimental_results}
\subsubsection{Long-Term Forecasting}
Table~\ref{tab:full-shot-comparison} reports long-term forecasting results on six datasets, comparing T2ST with full-shot task-specific model and the corresponding backbone models trained with 10\% or full ST data.

As shown in Table~\ref{tab:full-shot-comparison}, T2ST-trained OpenCity matches or outperforms full-data OpenCity in both MAE and RMSE on five of six datasets while using the 10\% ST budget. On PEMS07M, MAE decreases from 4.53 to 4.16 and RMSE from 8.30 to 7.81, corresponding to reductions of 8.17\% and 5.90\%, respectively. Both metrics also improve on CAD3, CAD5, and NYC-BIKE.

Under the same restricted ST budget, T2ST-trained UrbanFM reduces MAE on three datasets and RMSE on four compared with full-data UrbanFM. On PEMS07M, MAE decreases from 5.13 to 4.88 and RMSE from 9.93 to 9.31, yielding reductions of 4.87\% and 6.24\%, respectively. Its performance broadly comparable to that of the full-data model.

\subsubsection{Short-Term Forecasting}
Table~\ref{tab:urbanfm-10pct-5min} reports results on four grid-based datasets: TrafficSH, NYC-BIKE, NYC-TAXI, and Tdrive-BJ. With OpenCity as the backbone, the T2ST framework improves all three metrics over OpenCity across these datasets. Compared with full-data OpenCity, it achieves lower errors in nine of the twelve dataset--metric entries while using the 10\% ST budget. On TrafficSH, MAE decreases from 0.44 to 0.41 and RMSE from 0.97 to 0.88, corresponding to reductions of 6.82\% and 9.28\%, respectively.

With UrbanFM as the backbone, the T2ST framework matches or outperforms full-data UrbanFM in MAE on all four grid-based datasets under the same ST budget. On Tdrive-BJ, MAE and RMSE decrease by 14.94\% and 4.63\%, respectively, while MAPE falls from 33.92\% to 19.09\%.

On sensor-based datasets (Table~\ref{tab:urbanfm_t2st}), the T2ST framework achieves broadly comparable performance to the full-data models for both backbones, with lower errors in more than half of the comparisons, while using the 10\% ST budget.
\begin{table*}[!t]
    \centering
    \caption{Ablation of temporal-data composition for distillation with a fixed 10\% ST-data budget. The random setting replaces both with randomly sampled temporal data. Bold and underlined values indicate the best and second-best results, respectively.
    }
    \label{tab:distillation-data-ablation}

    \small
    \setlength{\tabcolsep}{3pt}
    \renewcommand{\arraystretch}{1.15}

    \begin{tabular*}{\textwidth}{
        @{\extracolsep{\fill}}l*{12}{c}@{}
    }
        \toprule
        \multirow{2}{*}{\textbf{Setting}}
        & \multicolumn{2}{c}{\textbf{CAD3}}
        & \multicolumn{2}{c}{\textbf{CAD5}}
        & \multicolumn{2}{c}{\textbf{PEMS07M}}
        & \multicolumn{2}{c}{\textbf{TrafficSH}}
        & \multicolumn{2}{c}{\textbf{CHI-TAXI}}
        & \multicolumn{2}{c}{\textbf{NYC-BIKE}} \\
        \cmidrule(lr){2-3}
        \cmidrule(lr){4-5}
        \cmidrule(lr){6-7}
        \cmidrule(lr){8-9}
        \cmidrule(lr){10-11}
        \cmidrule(lr){12-13}
        & MAE & RMSE
        & MAE & RMSE
        & MAE & RMSE
        & MAE & RMSE
        & MAE & RMSE
        & MAE & RMSE \\
        \midrule

        w/o $\mathcal{D}_{\mathrm{M}}$ and $\mathcal{D}_{\mathrm{D}}$
        & 17.07 & 28.85
        & 11.05 & 18.96
        & 4.22 & 7.84
        & \underline{0.56} & \textbf{0.85}
        & \textbf{2.33} & \underline{5.35}
        & \underline{6.09} & \underline{10.90} \\

        w/o $\mathcal{D}_{\mathrm{D}}$
        & \underline{16.52} & \underline{28.05}
        & \underline{10.81} & \underline{18.58}
        & \underline{4.19} & \textbf{7.79}
        & \underline{0.56} & \underline{0.86}
        & \underline{2.34} & \textbf{5.33}
        & \textbf{6.05} & \textbf{10.88} \\

        w/ random $\mathcal{D}_{\mathrm{M}}$ and $\mathcal{D}_{\mathrm{D}}$
        & 16.81 & 28.06
        & 11.17 & 19.03
        & 4.26 & 7.89
        & 0.59 & 0.89
        & 2.44 & 5.57
        & 6.47 & 11.36 \\

        \addlinespace[2pt]
        \textbf{w/ $\mathcal{D}_{\mathrm{M}}$ and $\mathcal{D}_{\mathrm{D}}$}
        & \textbf{16.25} & \textbf{27.72}
        & \textbf{10.74} & \textbf{18.51}
        & \textbf{4.16} & \underline{7.81}
        & \textbf{0.55} & \textbf{0.85}
        & \underline{2.34} & 5.41
        & \textbf{6.05} & 10.92 \\

        \bottomrule
    \end{tabular*}
\end{table*}

\begin{table}[t]
    \centering
    \caption{Forecasting performance of UrbanFM, OpenCity,
    and their T2ST variants. Full denotes the full ST training
    set; T2ST variants use the 10\% ST budget.
    }
    \label{tab:urbanfm_t2st}
    \footnotesize
    \setlength{\tabcolsep}{2pt}
    \renewcommand{\arraystretch}{1.08}

    \resizebox{\columnwidth}{!}{%
    \begin{tabular}{lcccccccc}
        \toprule
        \multirow{2}{*}{\textbf{Dataset}}
        & \multicolumn{2}{c}{\shortstack{\textbf{UrbanFM}\\\textbf{(Full)}}}
        & \multicolumn{2}{c}{\shortstack{\textbf{T2ST}\\\textbf{(UrbanFM)}}}
        & \multicolumn{2}{c}{\shortstack{\textbf{OpenCity}\\\textbf{(Full)}}}
        & \multicolumn{2}{c}{\shortstack{\textbf{T2ST}\\\textbf{(OpenCity)}}} \\
        \cmidrule(lr){2-3} \cmidrule(lr){4-5}
        \cmidrule(lr){6-7} \cmidrule(lr){8-9}
        & MAE & RMSE & MAE & RMSE & MAE & RMSE & MAE & RMSE \\
        \midrule
        OCC-Pairs   & 1.35 & 2.95 & 1.29 & 2.92 & 1.28 & 2.90 & 1.37 & 3.05 \\
        OCC-Hamburg & 6.23 & 10.38 & 6.20 & 10.41 & 6.19 & 10.71 & 6.11 & 10.45 \\
        PEMS-Bay    & 2.49 & 5.63 & 2.32 & 5.31 & 2.58 & 6.07 & 2.36 & 5.75 \\
        METR-LA     & 6.17 & 13.33 & 6.05 & 13.15 & 6.41 & 14.29 & 5.86 & 13.67 \\
        PEMS-03     & 21.53 & 34.33 & 21.40 & 34.50 & 21.61 & 35.39 & 22.25 & 34.93 \\
        PEMS-04     & 27.08 & 41.98 & 27.50 & 42.67 & 28.31 & 45.02 & 29.25 & 45.15 \\
        PEMS-07     & 31.52 & 47.93 & 31.95 & 48.53 & 32.49 & 51.45 & 32.97 & 50.46 \\
        PEMS-08     & 22.47 & 34.97 & 22.70 & 35.53 & 23.56 & 38.61 & 24.13 & 38.16 \\
        \bottomrule
    \end{tabular}%
    }
    \vspace{-4mm}
\end{table}

\subsection{Effect of Temporal Data Selection (RQ2)}
\label{sec:data_selection_results}

\subsubsection{Matched, Diverse, and Random Temporal Data}
Table~\ref{tab:distillation-data-ablation} compares four distillation settings under a fixed 10\% ST budget: using only ST data, adding randomly sampled temporal data $\mathcal{D}_{\mathrm{M}}$ and $\mathcal{D}_{\mathrm{D}}$, adding matched data $\mathcal{D}_{\mathrm{M}}$, and adding both matched and diverse data $\mathcal{D}_{\mathrm{M}}$ and $\mathcal{D}_{\mathrm{D}}$. Matched selection yields lower MAE and RMSE than equal-budget random sampling on all six datasets, supporting the selection of temporal patterns relevant to ST observations.

Adding diverse data $\mathcal{D}_{\mathrm{D}}$ further reduces MAE on four datasets and matches the matched-only result on the other two. The combined setting achieves  the best or tied-best MAE on five of six datasets and outperforms random sampling in every reported metric. On CAD3, MAE and RMSE reach 16.25 and 27.72, representing reductions of 4.80\% and 3.92\%, respectively, relative to distillation using only ST data.

\subsubsection{Choice of Selection Encoder}
Table~\ref{tab:encoder-ablation} replaces the frozen encoder used to construct temporal-data embeddings with TS2Vec, TS-TCC, or CoST. This changes the representation used for retrieval and clustering.

TS2Vec yields the lowest MAE on five of six datasets and the lowest RMSE on CAD3, CAD5, and TrafficSH. CoST performs best on CHI-TAXI for both metrics and gives the lowest RMSE on PEMS07M and NYC-BIKE. Across all six datasets, all three encoder configurations match or outperform the OpenCity baseline trained on the full training set. The gains are therefore observed with multiple selection encoders, although the most effective representation depends on the target dataset and metric. 
\begin{table}[!t]
\centering
\caption{Effect of different temporal encoders.}
\label{tab:encoder-ablation}
\small
\setlength{\tabcolsep}{3pt}
\renewcommand{\arraystretch}{1.15}
\resizebox{\columnwidth}{!}{%
\begin{tabular}{l*{6}{c}}
\toprule
\multirow{2}{*}{\textbf{Dataset}}
& \multicolumn{2}{c}{\textbf{TS2Vec}}
& \multicolumn{2}{c}{\textbf{TS-TCC}}
& \multicolumn{2}{c}{\textbf{CoST}} \\
\cmidrule(lr){2-3}
\cmidrule(lr){4-5}
\cmidrule(lr){6-7}
& MAE & RMSE
& MAE & RMSE
& MAE & RMSE \\
\midrule
CAD3
& 16.25 & 27.72
& 16.57 & 27.94
& 16.47 & 28.13 \\
CAD5
& 10.74 & 18.51
& 10.99 & 18.93
& 10.82 & 18.68 \\
PEMS07M
& 4.16 & 7.81
& 4.29 & 7.95
& 4.17 & 7.73 \\
TrafficSH
& 0.55 & 0.85
& 0.56 & 0.86
& 0.56 & 0.86 \\
CHI-TAXI
& 2.34 & 5.41
& 2.32 & 5.33
& 2.31 & 5.23 \\
NYC-BIKE
& 6.05 & 10.92
& 6.12 & 10.97
& 6.07 & 10.90 \\
\bottomrule
\end{tabular}%
}
\vspace{-4mm}
\end{table}

\subsubsection{Visualization of Temporal Patterns}
\label{sec:temporal_cluster_visualization}

\begin{figure}[!t]
    \centering
    \includegraphics[width=\columnwidth]{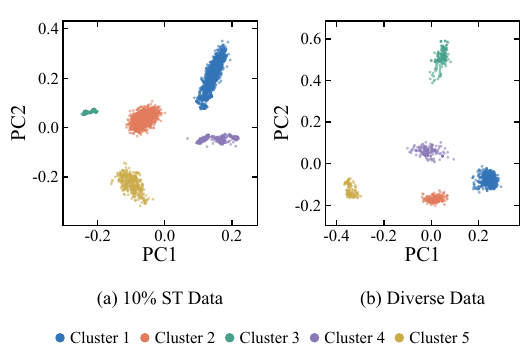}
    \caption{Five largest temporal clusters in (a) the 10\% ST subset and (b) the diverse data $\mathcal{D}_{\mathrm{D}}$.}
    \label{fig:temporal_clusters}
    \vspace{-6mm}
\end{figure}

Figure~\ref{fig:temporal_clusters} visualizes TS2Vec embeddings of temporal windows from the 10\% ST subset and the selected diverse data $\mathcal{D}_{\mathrm{D}}$. Each point represents a temporal window, with colors indicating the five largest HDBSCAN clusters in each panel. The clusters highlight recurring patterns in the ST observations and coherent temporal structure in the diverse data, providing a qualitative complement to the ablation results in Table~\ref{tab:distillation-data-ablation}.

\subsection{Effect of the Batch Mixing Ratio (RQ3)}
\label{sec:batch_mixing}

\begin{figure*}[!t]
    \centering
    \includegraphics[width=\linewidth]{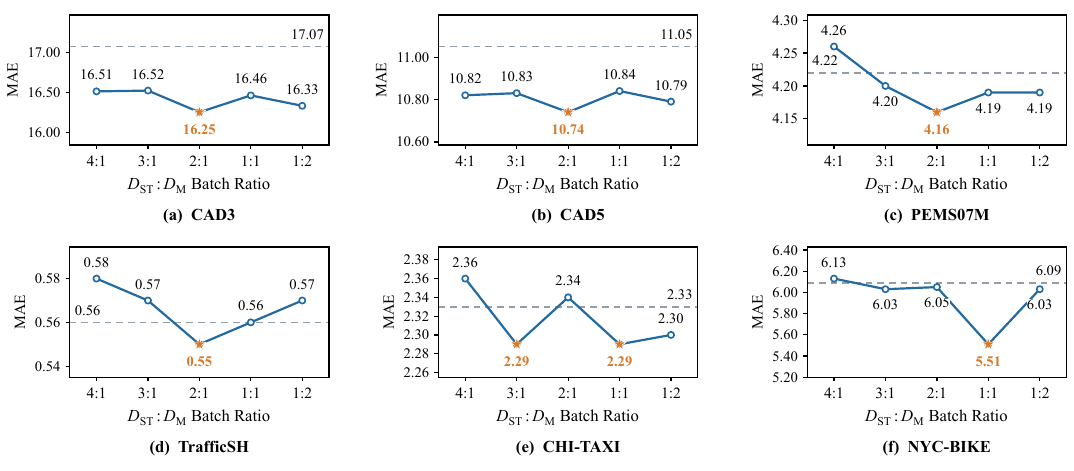}
    \caption{MAE under different ratios of ST to matched temporal batches. Horizontal dashed lines indicate the baseline for distillation using only ST data.}
    \label{st:matched ratio}
    \vspace{-4mm}
\end{figure*}

Figure~\ref{st:matched ratio} varies the scheduling ratio between ST and matched temporal batches, where $a:b$ schedules $b$ matched batches for every $a$ ST batches; ST-only distillation ($1:0$) provides the reference. Among the tested mixed ratios from $4:1$ to $1:2$, $2:1$ achieves the lowest MAE on CAD3, CAD5, PEMS07M, and TrafficSH. CHI-TAXI attains its lowest MAE at both $3:1$ and $1:1$, while NYC-BIKE favors $1:1$, indicating dataset-dependent scheduling preferences. We use $2:1$ as the common setting in the main experiments.

\begin{figure*}[!t]
    \centering
    \includegraphics[width=\linewidth]{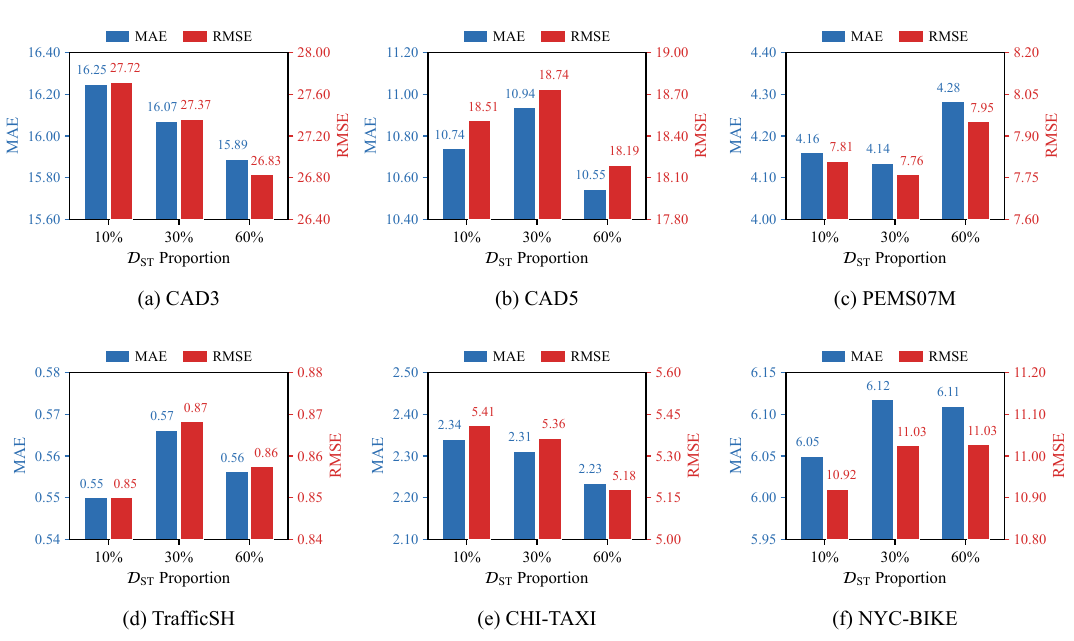}
    \caption{Forecasting performance of \textsc{T2ST} (OpenCity) with 10\%, 30\%, and 60\% ST-data budgets, each supplemented with auxiliary time-series data for knowledge distillation. The 60\% budget denotes the full training split.}
    \label{fig:st_ratio}
    \vspace{-4mm}
\end{figure*}

\subsection{Effect of the ST Data Budget (RQ4)}
\label{sec:st_budget}
Figure~\ref{fig:st_ratio} reports the performance of \textsc{T2ST} (OpenCity) after auxiliary time-series augmentation and knowledge distillation under different ST-data budgets. For each budget of 10\%, 30\%, or 60\%, we supplement the corresponding ST subset with auxiliary time-series data and distill knowledge from a frozen TSFM into the STFM student using both data types. These percentages refer only to the ST-data budget, with 60\% corresponding to the complete training split. Under this training procedure, increasing the ST budget does not consistently improve forecasting accuracy. Compared with the 30\% configuration, the 10\% configuration achieves lower MAE and RMSE on CAD5, TrafficSH, and NYC-BIKE. 

The comparison with the full 60\% training budget further supports the data efficiency of the 10\% setting. TrafficSH and NYC-BIKE attain their lowest MAE and RMSE at 10\%, while PEMS07M also performs better at 10\% than at 60\%. On CAD5, increasing the structured-data budget sixfold reduces MAE from 10.74 to 10.55 and RMSE from 18.51 to 18.19, corresponding to improvements of only 1.77\% and 1.73\%, respectively. Larger budgets benefit CAD3 and CHI-TAXI more consistently, but their MAE reductions from 10\% to 60\% remain approximately 2.22\% and 4.70\%. These results indicate that, with auxiliary time-series data and pretrained temporal knowledge incorporated through distillation, additional ST observations provide limited and dataset-dependent marginal gains. The 10\% configuration therefore offers a favorable accuracy--data-efficiency trade-off, using one-sixth of the full structured-data budget while achieving comparable or better accuracy on several datasets.

\subsection{Effect of Distillation Loss Weights (RQ5)}
\label{sec:distillation_weights}

We examine how the weights of temporal-relation distillation $\lambda_{\mathrm{rel}}$ and multi-scale prediction distillation $\lambda_{\mathrm{ms}}$ in Eq.~\eqref{eq:kd_batch_objective} affect the STFM under the 10\% ST budget. 

\begin{figure}[!t]
    \centering
    \includegraphics[width=\columnwidth]{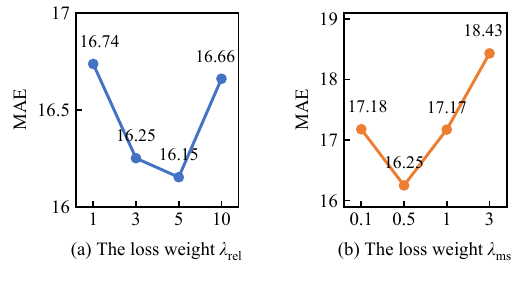}
    \caption{Sensitivity to distillation loss weights on CAD3.}
    \label{fig:lambda_ablation_cad3}
    \vspace{-5mm}

\end{figure}

Figure~\ref{fig:lambda_ablation_cad3} shows that, MAE on CAD3 first decreases and then increases as either distillation weight grows. With $\lambda_{\mathrm{ms}}$ fixed at $0.5$, the lowest MAE is $16.15$ at $\lambda_{\mathrm{rel}}=5$. With $\lambda_{\mathrm{rel}}$ fixed at $3$, the lowest MAE is $16.25$ at $\lambda_{\mathrm{ms}}=0.5$.

\section{Conclusion}
We presented T2ST, a data-efficient framework that combines selected auxiliary time series and knowledge from a frozen time-series foundation model to reduce STFM dependence on structured observations. Matched sequences reinforce observed patterns, while diverse sequences broaden temporal coverage through teacher reliability filtering. Temporal relation and multi-scale prediction distillation enhance the student’s shared temporal pathway, while real spatio-temporal samples provide spatial supervision.  Experiments with STFM on six long-term and twelve short-term datasets demonstrate competitive forecasting performance under the limited ST-data, matching or exceeding full-data counterparts on multiple datasets. These results support temporal-data selection and knowledge transfer as a practical approach to data-efficient STFM training.

\FloatBarrier
\balance
\bibliographystyle{IEEEtran}
\bibliography{references}

\end{document}